\documentclass[conference]{IEEEtran}
\IEEEoverridecommandlockouts
\usepackage{cite}
\usepackage{amsmath,amssymb,amsfonts}
\usepackage{algorithmic}
\usepackage{graphicx}
\usepackage{textcomp}
\usepackage{xcolor}

\usepackage{multirow}
\usepackage{array}
\usepackage{amsmath}

\usepackage{varwidth}
\usepackage{tikz}
\usepackage{subfigure}
\usepackage{pgfplots}
\pgfplotsset{compat=1.16}

\usepackage{amssymb}

\usepackage{booktabs}

\def\BibTeX{{\rm B\kern-.05em{\sc i\kern-.025em b}\kern-.08em
    T\kern-.1667em\lower.7ex\hbox{E}\kern-.125emX}}
\begin{document}

\title{
Reasoning Paths as Signals: Augmenting Multi-hop Fact Verification through Structural Reasoning Progression\\

{\footnotesize 
}
}

\author{
\IEEEauthorblockN{
Liwen Zheng\textsuperscript{1}, 
Chaozhuo Li\textsuperscript{1}, 
Haoran Jia\textsuperscript{1},
Xi Zhang\textsuperscript{1}
}
\IEEEauthorblockA{
\textsuperscript{1}Key Laboratory of Trustworthy Distributed Computing and Service (MoE) \\  Beijing University of Posts and Telecommunications   \\
\{zhenglw, lichaozhuo, jiahaoran, zhangx\}@bupt.edu.cn
}

}

\maketitle

\begin{abstract}
The growing complexity of factual claims in real-world scenarios presents significant challenges for automated fact verification systems, particularly in accurately aggregating and reasoning over multi-hop evidence. 
Existing approaches often rely on static or shallow models that fail to capture the evolving structure of reasoning paths, leading to fragmented retrieval and limited interpretability. 
To address these issues, we propose a Structural Reasoning framework for Multi-hop Fact Verification that explicitly models reasoning paths as structured graphs throughout both evidence retrieval and claim verification stages. 
Our method comprises two key modules: a structure-enhanced retrieval mechanism that constructs reasoning graphs to guide evidence collection, and a reasoning-path-guided verification module that incrementally builds subgraphs to represent evolving inference trajectories. 
We further incorporate a structure-aware reasoning mechanism that captures long-range dependencies across multi-hop evidence chains, enabling more precise verification.
Extensive experiments on the FEVER and HoVer datasets demonstrate that our approach consistently outperforms strong baselines, highlighting the effectiveness of reasoning-path modeling in enhancing retrieval precision and verification accuracy. 
\end{abstract}


\section{Introduction}
In the digital age, the unregulated spread of misinformation poses a profound threat to public understanding and societal stability. Automated fact verification has emerged as a critical tool to address this issue by evaluating whether a natural language claim aligns with established facts \cite{FEVER}. This task generally involves two core stages: retrieving relevant evidence from a large-scale knowledge corpus, and performing logical reasoning over the retrieved evidence to assess the claim's veracity \cite{reread}.
Effective fact verification not only enhances the reliability of online content but also plays a vital role in safeguarding public discourse and democratic processes.

\begin{figure}[t]
\centering
\includegraphics[width=1.0\linewidth]{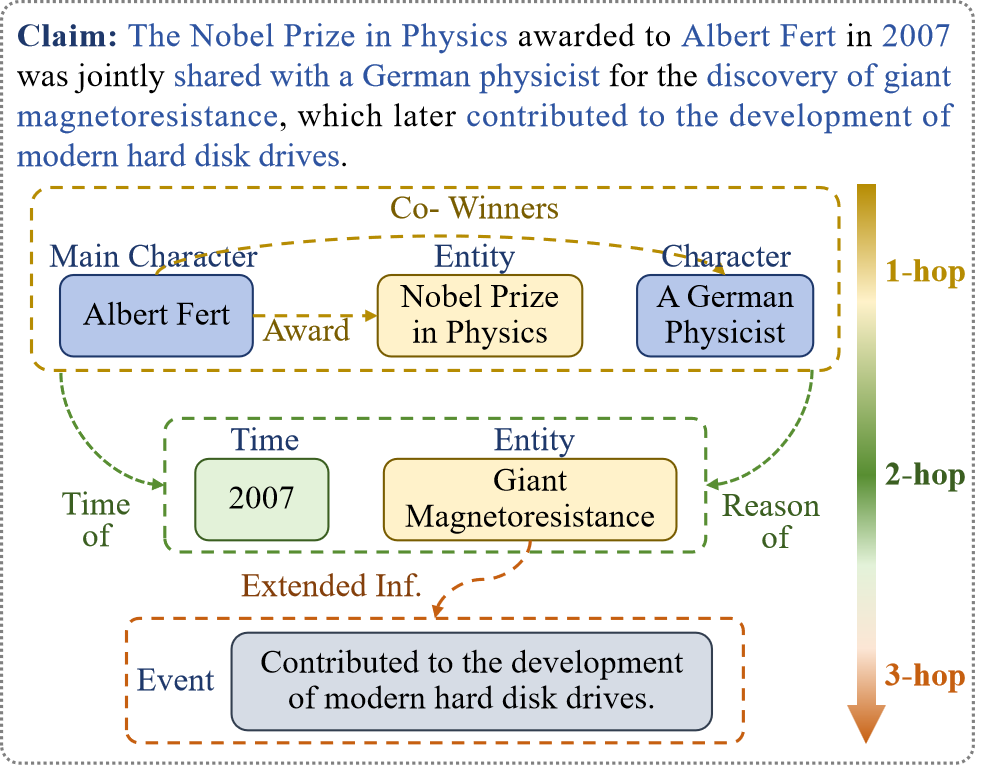}
\caption{A complex claim requiring multi-hop reasoning over entities, events, and relational dependencies.}
\label{intro}
\end{figure}

Conventional fact verification methods typically rely on the semantic similarity between a claim and candidate evidence sentences to retrieve a relevant evidence set, which is then used to assess the claim’s veracity\cite{KGAT, EvidenceNet}. While this approach performs well for simple, straightforward claims, it often fails when handling complex claims, which require reasoning over multiple entities, events, relational  dependencies, or causal relationships \cite{SAGP}.
As illustrated in Figure~\ref{intro}, verifying whether a scientist received a Nobel Prize for a discovery that later contributed to technological advancement involves complex reasoning. This includes connecting multiple entities, events, and causal relations—such as award attribution, scientific contribution, and downstream impact.
Traditional single-hop verification models lack mechanisms for linking information across evidence, often resulting in fragmented reasoning chains or missed critical evidence that render them ineffective for complex claims \cite{HESM}.
In contrast, multi-hop verification iteratively retrieves and integrates semantically linked evidence, offering a more robust framework for constructing coherent evidence sets and validating complex claims \cite{MLA}.

Multi-hop fact verification typically follows a stepwise retrieval process, where each query is generated from the claim and prior evidence to iteratively retrieve the next relevant piece, forming a complete set of evidence \cite{Politihop, MRR-FV}.
By modeling inter-evidence dependencies during retrieval, these methods have demonstrated improved capacity for handling complex claims.
Despite this progress, current multi-hop approaches face fundamental limitations. 
Firstly, in the retrieval stage, most systems generate the next-hop query by simply concatenating the claim with earlier evidence, without explicitly modeling the retrieving history \cite{Multi-Hop-DR}. 
This results in poorly defined reasoning paths that evolve from the original claim through step-by-step integration of evidence. 
Such unclear logical development makes it difficult to retrieve precise and context-aware information in later hops.
Secondly, in the verification stage, most approaches construct all evidence as a unified graph, ignoring the evolving nature of multi-hop reasoning  \cite{RMHR}. 
As a consequence, they fail to capture how evidence accumulates over time and construct interpretable reasoning results, ultimately limiting the transparency and accuracy of claim verification.

To address these limitations, we construct a multi-hop fact verification framework centered on explicit reasoning path modeling, designed to capture dependencies among evidence pieces to enhance both retrieval accuracy and reasoning coherence in complex claim verification. 
However, two major challenges must be overcome to realize this goal: 
(1) In the retrieval phase, effective retrieval requires the model to leverage the semantic and logical relationships among previously retrieved evidence to guide query generation for subsequent hops. 
Traditional approaches based on simple concatenation fail to capture the evolving logic among evidence, often causing the retrieval process to deviate from a coherent reasoning path.
(2) In the verification phase, the model should be capable of explicitly modeling the reasoning path, enabling it to highlight the progression of critical clues throughout multi-hop reasoning steps.
Yet current methods largely rely on static reasoning structures, which fail to reflect the progressive nature of multi-hop reasoning, making it difficult to handle long, dependency-rich evidence chains.

To address these challenges, we propose a \textbf{S}tructured \textbf{R}easoning framework for \textbf{M}ulti-hop \textbf{F}act \textbf{V}erification (SR-MFV) composed of two key modules: structure-enhanced multi-hop evidence retrieval and reasoning-path-guided claim verification. 
Together, these modules explicitly model the evolving reasoning path during both retrieval and verification phases, enabling more effective verification of complex claims.
In the evidence retrieval module, the original claim is integrated with previously retrieved evidence into a graph structure that captures semantic, entity-level, and event-based dependencies among evidence pieces. 
This reasoning graph provides structured contextual signals to guide the generation of the next-hop query, ensuring that subsequent retrieval remains aligned with the reasoning trajectory and minimizing off-path or redundant evidence acquisition.
The claim verification module adopts a progressive evidence graph construction and reasoning mechanism to model the evolving structure of the reasoning path.
A sequence of subgraphs are incrementally constructed in the order of evidence retrieval, explicitly recording the gradual accumulation and logical evolution of evidence.
To support deeper multi-hop reasoning, we incorporate GraphFormers \cite{GraphFormers} to process these evidence subgraphs. 
GraphFormers integrate structure-aware attention biases with the global receptive field of Transformers, enabling the modeling of long-term dependencies between non-adjacent nodes in a graph.
The final prediction is obtained by aggregating reasoning representations across the full sequence of subgraphs. 
Overall, this method captures both local inference dynamics and global path consistency, thereby enhancing the accuracy of complex claim verification.

Our main contributions can be summarized as follows:
\begin{itemize}

    \item This paper proposes a structural reasoning framework for multi-hop fact verification that explicitly models the reasoning path across multiple evidence pieces, overcoming the fragmentation limitations of traditional approaches and effectively addressing the challenges of complex claim verification.
    \item The proposed SR-MFV model enhances verification performance through the collaboration of two key components:
    The structure-enhanced multi-hop evidence retrieval module guides context-aware retrieval at each hop via structure-enhanced queries, facilitating the construction of a coherent reasoning path. 
    The reasoning-path-guided claim verification module explicitly models the reasoning path through progressive subgraph construction and reasoning, thereby strengthening the extraction of critical clues. 
    \item Extensive experiments on both general and multi-hop fact verification datasets demonstrate the model’s superiority in handling complex claims, showing clear improvements in both retrieval precision and verification accuracy.
\end{itemize}

\section{Related Work}
Early efforts in automated fact verification predominantly adopt single-hop frameworks, retrieving evidence based on semantic similarity to the claim and verifying in a single step \cite{SURVEY}.
Classical methods treat retrieval as an isolated matching task \cite{RAV}, for example, KGAT \cite{KGAT} enhances ESIM for fine-grained alignment, Deformer \cite{DeFormer} scores document pairs via Siamese networks, and LIST5 \cite{ListT5} employs pre-trained models for evidence ranking.
Although effective, these dense retrieval approaches often incur high computational costs and limited efficiency.
To address this, generation-based retrieval has emerged, leveraging autoregressive models to directly generate relevant entities or passages \cite{AER}.
GERE \cite{GERE} exemplifies this shift, replacing expensive ranking with lightweight generation.
Nevertheless, single-hop systems inherently lack the capacity to capture inter-evidence dependencies, thus falling short in complex scenarios that require reasoning across multiple clues \cite{Politihop}.

To overcome the limitations of single-hop approaches, multi-hop fact verification frameworks have been developed to aggregate and reason over interrelated pieces of evidence, explicitly modeling reasoning paths to capture key inferential clues~\cite{HOVER, Self-RAG}.
In the retrieval phase, HESM~\cite{HESM} introduces a hierarchical multi-hop mechanism to construct evidence sets. However, its heavy reliance on extensive entity linking restricts its applicability in low-resource or sparsely annotated datasets.
An alternative paradigm is dense multi-hop retrieval~\cite{Multi-Hop-DR}, where the original query is progressively augmented with previously retrieved evidence to facilitate subsequent hops~\cite{RMHR, RECOMP}.
While these methods consider inter-evidence interactions during retrieval, most rely solely on semantic concatenation for query enhancement—an approach insufficient for capturing deeper logical dependencies across evidence. As a result, constructing faithful and coherent evidence chains remains a challenge~\cite{faithful-rationale}.
In the verification phase, many systems represent multi-hop evidence as a unified reasoning graph and apply graph-based inference to extract key clues \cite{LOREN}. Yet, the reasoning capabilities of conventional models are often inadequate for capturing the complex semantic and causal relationships required for accurate claim verification~\cite{Causal-Walk}.

\section{Problem Definition}
The fact verification task involves assessing the veracity of a given claim $c$ based on information retrieved from a large evidence corpus $\mathcal{D}$. The objective is to categorize the claim into one of three labels: \textsc{Supported}, \textsc{Refuted}, or \textsc{Not Enough Information}, by retrieving and reasoning over a sequence of evidence sentences $E = \{e_1, e_2, ..., e_n\} \subset \mathcal{D}$ that collectively support or refute the claim through a coherent reasoning path.
Unlike single-hop verification, 
complex claims typically require integrating multiple pieces of evidence that span different documents, entities, or semantic relations. This necessitates a multi-hop process capable of constructing intermediate connections and logical dependencies across evidence.
This multi-hop solution is composed of two key components:

\begin{itemize}
    \item \textbf{Multi-hop Evidence Retrieval,} where relevant evidence is selected in multiple steps. At the $t$-th retrieval step, a query is generated based on the original claim $c$ and the previously retrieved evidence $\{e_1, ..., e_{t-1}\}$, which is used to retrieve the next evidence $e_t$ from the corpus.
    
    \item \textbf{Claim Verification,} where the collected evidence is jointly reasoned over to infer the final veracity label $y$.

\end{itemize}

\section{Methodology}
\begin{figure*}[t]
\centering
\includegraphics[width=0.85
\linewidth]{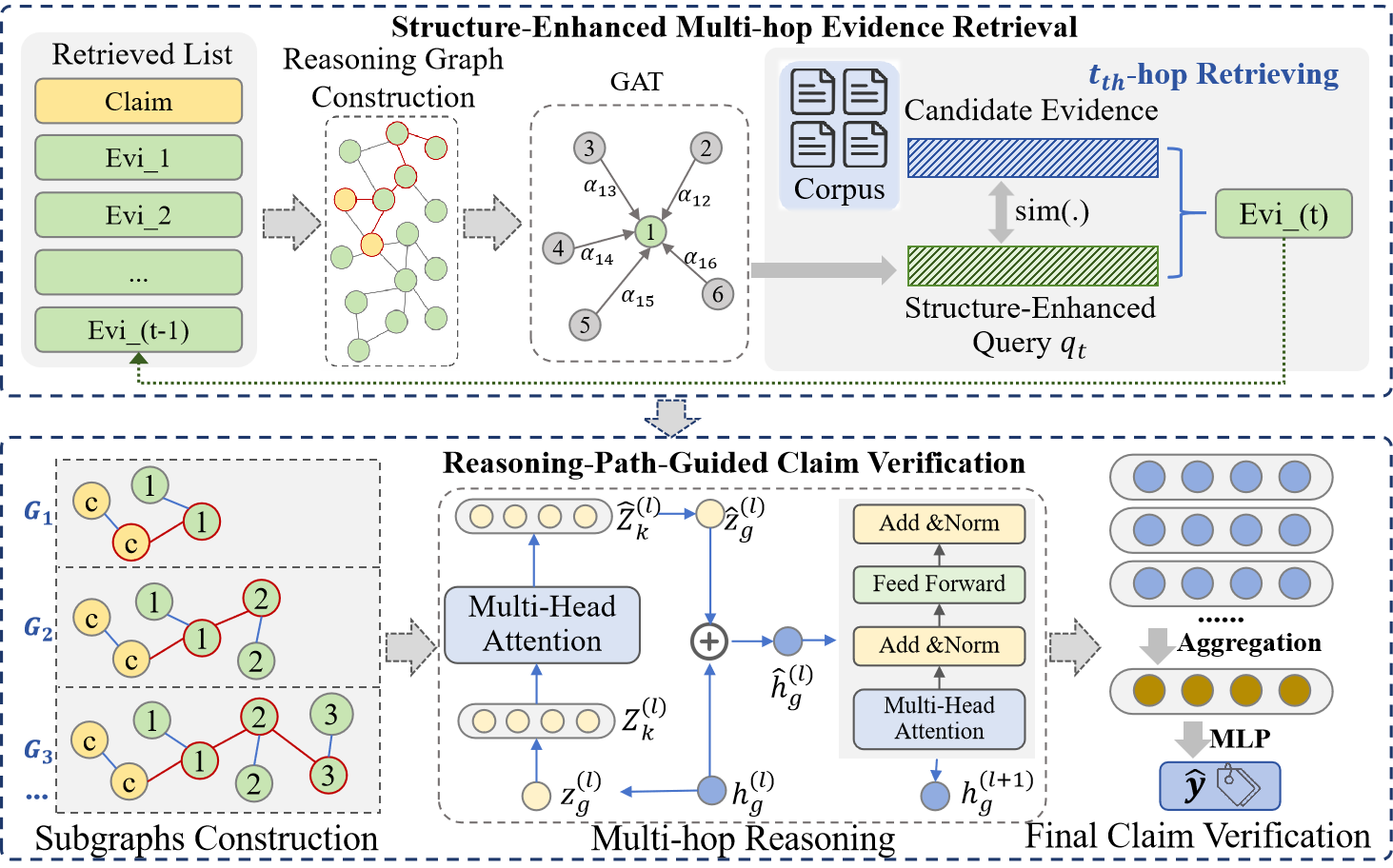}
\caption{Framework of the proposed SR-MFV model.}
\label{model}
\end{figure*}

As depicted in Figure~\ref{model}, the proposed SR-MFV model consists of two main components: (1) the Structure-Enhanced Multi-hop Evidence Retrieval module that incrementally builds reasoning paths, and (2) the Reasoning-Path-guided Claim Verification module that performs progressive inference over evidence subgraphs.

\subsection{Structure-Enhanced Multi-hop Evidence Retrieval}
A major limitation of existing multi-hop retrieval approaches lies in their reliance on simple textual concatenation of the claim and previously retrieved evidence to form the next-hop query. 
This strategy treats prior evidence as unstructured context, failing to capture the underlying dependencies between evidence pieces. 
Consequently, such methods are prone to retrieval drift, redundant results, or fragmented reasoning paths.

To address these limitations, we introduce a structure-enhanced retrieval module that models the original claim and previously retrieved evidence as a unified reasoning graph. 
At the $t$-th retrieval step, we construct a reasoning graph $G_{t-1} = (V, E)$, where the node set $V$ consists of all tokens from the claim $c$ and the retrieved evidence $\{e_1, e_2, \dots, e_{t-1}\}$. 
Edges in $E$ are defined based on intra-sentence adjacency and inter-sentence co-occurrence. 
This graph formulation enables fine-grained modeling of dependencies across evidence and allows the retrieval query to reflect not only sentence-level semantics but also the evidence interactions within the evolving reasoning path.

We perform graph attention-based message passing over the constructed reasoning graph to derive a unified representation that guides the retrieval of the next evidence sentence. Each node $v_i \in V$ is first encoded using a pre-trained language model to capture contextual semantics:
\begin{equation}
\mathbf{h}_i^{(0)} = \text{Enc}(v_i), \quad \forall v_i \in V .
\end{equation}

To incorporate structural context, we apply $L$ layers of graph attention propagation. At each layer $l$, the hidden representation of node $v_i$ is updated by attending to its neighbors:
\begin{equation}
\mathbf{h}_i^{(l)} = \sigma \left( \sum_{j \in \mathcal{N}(i)} 
\alpha_{ij}^{(l)} \mathbf{W}_{l} \mathbf{h}_j^{(l-1)} \right),
\end{equation}
where $\mathcal{N}(i)$ denotes the set of neighboring nodes of $v_i$ in $G_{t-1}$, $\mathbf{W}_{l}$ is a trainable projection matrix at layer $l$, and $\sigma(\cdot)$ is a non-linear activation function (e.g., ReLU). The attention coefficient $\alpha_{ij}^{(l)}$ measures the importance of node $v_j$ to $v_i$ and is computed as:
\begin{equation}
\alpha_{ij}^{(l)} = \frac{ \exp \left( \phi \left( \mathbf{a}^\top \left[ \mathbf{W}_{l} \mathbf{h}_i^{(l-1)} \, \Vert \, \mathbf{W}_{l} \mathbf{h}_j^{(l-1)} \right] \right) \right) }{ \sum_{k \in \mathcal{N}(i)} \exp \left( \phi \left( \mathbf{a}^\top \left[ \mathbf{W}_{l} \mathbf{h}_i^{(l-1)} \, \Vert \, \mathbf{W}_{l} \mathbf{h}_k^{(l-1)} \right] \right) \right) }.
\end{equation}
Here, $\mathbf{a}$ is a learnable attention vector and $\Vert$ denotes vector concatenation. The function $\phi(\cdot)$ denotes the LeakyReLU activation. This attention mechanism allows each node to selectively aggregate information from its neighbors based on semantic and structural relevance.

After $L$ layers of propagation, the structure-aware token representations $\{\mathbf{h}_i^{(L)}\}$ are subsequently aggregated to derive the retrieval query vector $q_t$, which encodes both contextual semantics and graph-informed relations.
\begin{equation}
q_t = \sum_{i \in V} \beta_i \cdot \mathbf{h}_i^{(L)}, \quad \beta_i = \text{softmax}\left( \mathbf{w}^\top \tanh(\mathbf{W}_q \mathbf{h}_i^{(L)}) \right),
\end{equation}
where $\mathbf{W}_q$ and $\mathbf{w}$ are trainable parameters. This attention-based readout allows the model to emphasize key tokens (e.g., salient entities or events) within the reasoning graph when generating the next-hop retrieval query.

The resulting structure-aware query $q_t$ is then used to retrieve the next evidence sentence $e_t$ from the corpus based on semantic similarity:
\begin{equation}
e_t = \arg\max_{s \in \mathcal{D}} \, \text{sim}(q_t, \text{Enc}(s)),
\end{equation}
where $\text{Enc}(s)$ is the semantic embedding of candidate evidence $s$, and $\text{sim}(\cdot,\cdot)$ denotes a similarity function. 

By grounding the retrieval process in a semantically and structurally enriched reasoning graph, the model captures cross-evidence dependencies more effectively, enabling the construction of logically coherent reasoning paths.

\subsection{Reasoning-Path-Guided Claim Verification}

While the retrieval module focuses on progressively constructing a reasoning path, the verification module leverages its structure for fine-grained inference. 
Instead of processing the final evidence set as a static input, we explicitly encode the stepwise evolution of reasoning by constructing a sequence of evidence subgraphs. 
Each subgraph reflects an intermediate reasoning state along the multi-hop path, enabling the model to capture how critical information emerges and accumulates during the verification process.  
To this end, this module consists of three stages: (1) subgraphs construction, (2) multi-hop reasoning, and (3) final claim verification.

\subsubsection{Subgraphs Construction}
Given the claim $c$ and the sequence of retrieved evidence $E = \{e_1, e_2, \dots, e_n\}$, we construct a series of progressively growing evidence subgraphs $\{G_1, G_2, \dots, G_n\}$ to explicitly capture the evolving nature of the reasoning path. Each subgraph $G_k = (V_k, E_k)$ is constructed from the claim and the first $k$ evidence sentences $\{e_1, \dots, e_k\}$, preserving the retrieval order as a proxy for the logical inference process.

The node set $V_k$ consists of all tokens contained in $c$ and $\{e_1, \dots, e_k\}$, and the edge set $E_k$ is constructed using multiple types of relations: 
(1) \textit{Intra-sentence adjacency edges} ($E_k^{\text{adj}}$), which connect adjacent tokens within the same sentence to preserve local syntactic continuity; 
(2) \textit{Inter-sentence co-reference edges} ($E_k^{\text{coref}}$), which link coreferent mentions across different sentences, thereby facilitating coherent inference over distributed evidence; and 
(3) \textit{Learned latent edges} ($E_k^{\text{sem}}$), which are dynamically added through graph structure learning based on pairwise node representations, allowing the model to capture latent relations beyond explicit linguistic cues.

To support the construction of $E_k^{\text{sem}}$, we dynamically infer additional edges beyond explicit linguistic cues by computing pairwise affinities $w_{ij}$ between nodes:
\begin{equation}
w_{ij} = \text{sim}(\mathbf{h}_i, \mathbf{h}_j) = \mathbf{h}_i^\top \mathbf{W}_s \mathbf{h}_j,
\end{equation}
where $\mathbf{h}_i, \mathbf{h}_j \in \mathbb{R}^d$ are contextualized embeddings of nodes $v_i$ and $v_j$, and $\mathbf{W}_s \in \mathbb{R}^{d \times d}$ is a trainable similarity matrix.

To determine whether a learned edge should be retained in $E_k^{\text{sem}}$, we apply a sigmoid transformation to the similarity scores, followed by thresholding to select semantically meaningful connections:
\begin{equation}
E_k^{\text{sem}} = \left\{ (v_i, v_j) \mid \sigma(w_{ij}) > \tau \right\},
\end{equation}
where $\sigma(\cdot)$ is the sigmoid activation and $\tau$ is a tunable threshold. This threshold-based selection enables the model to flexibly retain latent informative semantic links, facilitating the capture of long-range dependencies while alleviating topological sparsity in the token-level reasoning graph. 

The final edge set is then composed by unifying the learned edges with linguistically grounded structures:
\begin{equation}
E_k = E_k^{\text{adj}} \cup E_k^{\text{coref}} \cup E_k^{\text{sem}}.
\end{equation}

Building on the above graph construction, we obtain a relationally enriched evidence graph that supports expressive multi-hop inference across both local and global contextual cues. 
By incrementally incorporating retrieved evidence, we construct a sequence of structured subgraphs $\{G_1, G_2, \dots, G_n\}$, each representing a distinct reasoning state along the reasoning path. 
Compared to static graph modeling, this progressive formulation offers finer control over reasoning dynamics and enables more interpretable and context-sensitive verification.

\subsubsection{Multi-hop Reasoning}
After constructing the subgraph sequence $\{G_1, G_2, \dots, G_n\}$, we perform multi-hop reasoning over each graph to extract informative clues for downstream tasks. 
Traditional GNNs, however, rely on shallow, local message passing, limiting their capacity to capture deep semantic interactions and long-range structural dependencies essential for complex reasoning.
To overcome these limitations, we adopt GraphFormers—a Transformer-based architecture that interleaves token-level encoding with graph-structured aggregation at each layer. 
This unified design enables joint reasoning over textual content and relational structure, enhancing the model’s ability to capture complex dependencies across multi-hop evidence graphs.

Formally, we apply GraphFormers independently to each evidence subgraph $G_k = (V_k, E_k)$ to capture both local semantics and relational structure. Let $\mathbf{h}_g^{(0)}$ denote the initial representations of node $v_g \in V_k$, formed by summing token and positional embeddings.
At each subsequent layer $l = 1, \dots, L$, GraphFormers perform a two-stage nested operation that interleaves self-attention over token sequences with structure-aware message passing across the graph.

\noindent\textbf{Graph Aggregation:} At each layer $l$, we extract a representation $\mathbf{z}_g^{(l)}$ by reading from the [CLS] token of $\mathbf{h}_g^{(l)}$, which serves as the global summary of the node $v_g$:
\begin{equation}
\mathbf{z}_g^{(l)} = \mathbf{h}_g^{(l)}[0].
\end{equation}

To enable global context integration over the subgraph structure, we first gather all node representations at layer $l$ into a matrix $\mathbf{Z}_k^{(l)} = [\mathbf{z}_g^{(l)}]_{g \in V_k}$. The resulting embedding matrix is then passed to a Multi-Head Attention (MHA) layer, which performs relation-aware aggregation over the subgraph tokens:
\begin{equation}
\hat{\mathbf{Z}}_k^{(l)} = \text{MHA}(\mathbf{Z}_k^{(l)}) = \text{Concat}(\text{head}_1, \dots, \text{head}_h),
\end{equation}
\begin{equation}
\text{head}_j = \text{softmax}\left( \frac{\mathbf{Q}_j \mathbf{K}_j^\top + \mathbf{B}}{\sqrt{d}} \right) \mathbf{V}_j,
\end{equation}
where $\mathbf{Q}_j = \mathbf{Z}_k^{(l)} \mathbf{W}_j^Q$, $\mathbf{K}_j = \mathbf{Z}_k^{(l)} \mathbf{W}_j^K$, and $\mathbf{V}_j = \mathbf{Z}_k^{(l)} \mathbf{W}_j^V$ are the query, key, and value projections for head $j$, and $\mathbf{B}$ is a learnable bias encoding graph topology.

Each updated node embedding $\hat{\mathbf{z}}_g^{(l)}$ is re-injected into the original token sequence as a prepended [CLS] vector, yielding a structure-augmented node representation:
\begin{equation}
\hat{\mathbf{h}}_g^{(l)} = [\hat{\mathbf{z}}_g^{(l)}] \, \Vert \, \mathbf{h}_g^{(l)}.
\end{equation}

\noindent\textbf{Textual Encoding:} To further integrate graph-level context into node representations, we apply an asymmetric Transformer attention mechanism:
\begin{equation}
\mathbf{h}_g^{(l+1)} = \text{Transformer}^{(l)}(\hat{\mathbf{h}}_g^{(l)}),
\end{equation}
where only the augmented [CLS] token attends to the full graph-enhanced sequence, enabling directional integration of structural cues.

Following $L$ layers of reasoning-aware encoding, we extract the final [CLS] embedding for each node $v_g \in V_k$, which serves as a condensed summary of its graph-informed context. These node-level summaries are then aggregated to form the full subgraph representation:
\begin{equation}
\mathbf{h}_g = \mathbf{h}_g^{(L)}[0], \quad 
\mathbf{H}_k = [\mathbf{h}_g]_{g \in V_k}.
\end{equation}

This nested architecture allows GraphFormers to perform joint text-structure reasoning at every layer, enabling deeper, more global context integration than traditional cascaded Transformer+GNN pipelines. In the context of multi-hop fact verification, this layered interaction is critical for capturing subtle dependencies between the claim and distributed evidence, providing a discriminative representation for downstream verification.

\subsubsection{Final Claim Verification} 
Once the sequence of subgraph representations $\{\mathbf{H}_1, \mathbf{H}_2, \dots, \mathbf{H}_n\}$ is derived from the multi-step reasoning process, we integrate them into a single, unified representation to facilitate final claim verification. Each $\mathbf{H}_k$ encodes the contextual understanding accumulated after incorporating the $k$-th piece of evidence. As such, uniformly aggregating all subgraphs risks diluting the differential relevance of each reasoning stage. To capture this nuance, we implement an attention-based fusion strategy that adaptively modulates the influence of each subgraph:
\begin{equation}
\alpha_k = \frac{\exp(\mathbf{w}^\top \tanh(\mathbf{W} \mathbf{H}_k))}{\sum_{j=1}^{n} \exp(\mathbf{w}^\top \tanh(\mathbf{W} \mathbf{H}_j))}, 
\quad 
\end{equation}
\begin{equation}
\mathbf{H} = \sum_{k=1}^{n} \alpha_k \cdot \mathbf{H}_k,
\end{equation}
where $\mathbf{W} $ and $\mathbf{w}$ are learnable parameters for attention scoring, and $\alpha_k$ denotes the normalized importance of subgraph $G_k$. 
In the end, the final representation $\mathbf{H}$ is passed through a multi-layer perceptron (MLP) classifier to predict the veracity label $\hat{y} \in \{\textsc{Supported}, \textsc{Refuted}, \textsc{NotEnoughInfo}\}$.


\section{Experiments}

\subsection{Experimental Setup}
This section describes the datasets, evaluation metrics, and baselines of the experiments. 

\noindent \textbf{Datasets}
We evaluate SR-MFV on two widely used fact verification datasets: FEVER \cite{FEVER} and HoVer \cite{HOVER}.
FEVER is a large-scale dataset containing 185,445 claims generated from Wikipedia. Each claim is labeled as \textsc{Supported}, \textsc{Refuted}, or \textsc{NotEnoughInfo}, and accompanied by one or more evidence sentences required to verify it. 
HoVer is a more challenging benchmark specifically designed for multi-hop fact verification. It consists of more than 26,000 claims annotated with fine-grained multi-hop evidence chains across multiple Wikipedia documents.

\noindent \textbf{Evaluation Metrics}
Following prior work \cite{CO-GAT}, we adopt different evaluation metrics for claim verification and evidence retrieval across the two datasets.
For the FEVER dataset, we use Label Accuracy (LA) and the official FEVER Score to evaluate claim verification performance. The FEVER Score measures the proportion of claims that are not only correctly classified but also supported by the gold evidence. For evidence retrieval, we report Precision, Recall, and F1, comparing the retrieved evidence set against the annotated gold evidence.
For the HoVer dataset, we follow its standard protocol and use Label Accuracy (LA) to assess claim verification. For evidence retrieval, we report F1 score based on whether the gold supporting sentences are retrieved, considering accuracy of multi-hop evidence chains.

\noindent \textbf{Baselines} 
Since most claims in the FEVER dataset can be resolved with one or two evidence sentences, we follow prior work and compare our model against a range of representative one-hop baselines. These include models such as KGAT \cite{KGAT}, DREAM \cite{DREAM}, EvidenceNet \cite{EvidenceNet}, and GEAR \cite{GEAR}, which typically rely on shallow sentence-level reasoning without modeling multi-hop dependencies. In addition, we include two multi-hop baselines: HESM \cite{HESM}, which leverages hyperlink structures across documents to guide evidence integration, and CO-GAT \cite{CO-GAT}, which constructs a unified evidence graph and applies GAT-based reasoning over the full retrieved context. Notably, models such as KGAT, DREAM, EvidenceNet, and CO-GAT perform global reasoning over flattened evidence graphs, but do not explicitly model the progression of the reasoning path.

For the more challenging HoVer benchmark, we compare our model to both one-hop methods (e.g., DeBERTa \cite{DeBERTa}, GEAR, and EvidenceNet) and multi-hop baselines (e.g., CO-GAT and SAGP \cite{SAGP}) to assess its ability to reason over long, cross-document evidence chains.

\subsection{Overall Performance}
As shown in Table~\ref{fever-overall} and Table~\ref{hover-overall}, our proposed SR-MFV model achieves the best overall performance in claim verification across both datasets, outperforming all compared baselines. 
This validates the effectiveness of explicitly modeling reasoning paths throughout both retrieval and verification stages.
While multi-hop models are inherently designed to handle complex reasoning, their performance advantage over one-hop baselines on the FEVER dataset remains relatively limited. This can be attributed to the nature of FEVER claims, which are generally straightforward and often solvable using a single sentence or a shallow combination of two. As a result, the full potential of multi-hop reasoning is not fully exploited in this setting.
In contrast, the HoVer dataset poses a more challenging scenario, with claims requiring longer reasoning chains and evidence distributed across multiple documents. In this context, multi-hop models demonstrate a clearer advantage, highlighting the necessity of structured multi-step inference for complex claims.

To further examine SR-MFV’s robustness on complex cases, we conduct a breakdown analysis on HoVer by grouping claims based on the number of required evidence hops, as annotated in the dataset. SR-MFV achieves consistent gains of 1.48\%, 1.51\%, and 3.09\% in label accuracy on 2-hop, 3-hop, and 4-hop subsets respectively, compared to the strongest baseline. 
These results highlight that MFV's ability to build and reason over evolving reasoning paths is particularly beneficial in scenarios where the complexity of claims demands fine-grained evidence coordination and structural reasoning.

\begin{table}[t]
\caption{Overall performance on FEVER. \textbf{Bold} indicates the best result, while \underline{underline} denotes the second best.}
\begin{center}
\begin{small}
\renewcommand{\arraystretch}{1.2}
\resizebox{0.85\columnwidth}{!}{
\begin{tabular}{lcccc}
\toprule
         & \multicolumn{2}{c}{\textbf{Dev}} & \multicolumn{2}{c}{\textbf{Test}} \\ \cmidrule(r){2-3} \cmidrule(r){4-5}
\textbf{Models} & \textbf{LA}          & \textbf{FEVER}        & \textbf{LA}           & \textbf{FEVER}          \\ 

\midrule
BERT Concat \cite{KGAT}  & 73.67  & 68.89  & 71.01  & 65.64  \\
GAT \cite{KGAT}  & 76.13  & 71.04  & 72.03  & 67.56  \\
GEAR \cite{GEAR}  & 74.84  & 70.69 & 71.60  &  67.10 \\
KGAT \cite{KGAT}  & 78.29  & 76.11 & 74.07  & 70.38  \\
DREAM \cite{DREAM}  & 79.16  & - & 76.85  & 70.60  \\
KGAT+GERE \cite{GERE}  & 79.44  & 77.38 &  75.24  &  71.17  \\
TARSA \cite{TARSA}  & 81.24  &  77.96 & 73.97  & 70.70  \\
Proofver \cite{Proofver}  & 80.74  & 79.07 & \underline{79.47}  & \underline{76.82} \\
EvidenceNet \cite{EvidenceNet}  & 81.46  &  78.29 &  76.95  & 73.78  \\

\midrule
HESM \cite{HESM}  & 75.77  & 73.44 & 74.64  &  71.48  \\
CO-GAT \cite{CO-GAT}  & \underline{81.56}  &  \underline{79.21} & 76.95  &  73.48  \\

\midrule
 \textbf{SR-MFV}  & \textbf{82.44}        & \textbf{79.67}        & \textbf{80.17}             & \textbf{77.62} \\

\bottomrule
\end{tabular}}

\label{fever-overall}
\end{small}
\end{center}
\end{table}

\begin{table}[t]
\caption{Overall performance on HOVER.}
\begin{center}
\begin{small}
\resizebox{0.8\columnwidth}{!}{
\begin{tabular}{llccc}
\toprule
\textbf{ } & \textbf{Models} & \textbf{2-hop} & \textbf{3-hop} & \textbf{4-hop} \\ 
\midrule

\multirow{3}{*}{\textbf{One-hop}} 
& DeBERTa \cite{DeBERTa} & 72.94 & 71.67 & 70.34 \\
& GEAR \cite{GEAR} & 73.50 & 72.33 & 71.79 \\
& EvidenceNet \cite{EvidenceNet} & 73.95 & 73.23 & 72.46 \\
\midrule

\multirow{2}{*}{\textbf{Multi-hop}} 
& CO-GAT \cite{CO-GAT} & 77.85 & 76.40 & 75.11 \\
& SAGP \cite{SAGP} & \underline{77.90} & \underline{76.78} & \underline{76.01} \\
\midrule

& \textbf{SR-MFV} & \textbf{79.05}   & \textbf{77.93}   & \textbf{78.36} \\

\bottomrule
\end{tabular}}

\label{hover-overall}
\end{small}
\end{center}
\end{table}

\subsection{Performance on Evidence Retrieval}
For the FEVER dataset, we follow the experimental setup of CO-GAT to evaluate evidence retrieval performance. As shown in Table~\ref{evi-fever}, although SE-MFV yields slightly lower recall compared to CO-GAT, it achieves significantly higher precision and overall F1. This improvement likely stems from SR-MFV’s structure-enhanced query generation, which guides retrieval along coherent reasoning paths. 
By focusing on semantically and relationally aligned evidence, it retrieves fewer but more precise sentences, improving precision and F1. 
The slightly lower recall is a result of this conservative strategy, where marginally relevant evidence may be omitted to preserve path relevance.

To further assess SR-MFV’s retrieval performance on complex claims, we compare its results with the strongest baseline across different levels of reasoning complexity on the \textsc{HoVer} dataset. As depicted in Table~\ref{evi-hover}, SR-MFV achieves consistent F1 improvements of 0.49\%, 0.73\%, and 1.12\% on the 2-hop, 3-hop, and 4-hop subsets, 
respectively. This mirrors the trend observed in overall verification accuracy and further demonstrates MFV’s strength in capturing multi-hop semantic dependencies during evidence retrieval.

\begin{table}[t]
\caption{Retrieval performance comparison on  FEVER(Dev).} 
\begin{center}
\begin{small}
\resizebox{0.7\columnwidth}{!}{
\begin{tabular}{lccc}
\toprule

 \textbf{Models} & \textbf{Prec@5} & \textbf{Rec@5} & \textbf{F1@5}\\ 

\midrule
UNCNLP \cite{UNCNLP}   &36.49 &86.79 &51.38   \\
DREAM \cite{DREAM} & 26.67 &87.64 &40.90 \\
MLA \cite{MLA}  & 25.63 &88.64 &39.76  \\
ICMI \cite{ICMI} & 25.74 &92.86 &40.30\\
GEAR \cite{GEAR} & \underline{40.60} &86.36 &\underline{55.23}  \\
CO-GAT \cite{CO-GAT} & 27.29 &\textbf{94.37} &42.34\\

\midrule
\textbf{SR-MFV}  & \textbf{43.73}   & \underline{90.34}    & \textbf{58.93}\\

\bottomrule
\end{tabular}}

\label{evi-fever}
\end{small}
\end{center}
\end{table}

\begin{table}[t]
\caption{Retrieval performance comparison on HOVER.} 
\begin{center}
\begin{small}
\resizebox{0.9\columnwidth}{!}{
\begin{tabular}{llcccc}
\toprule
\textbf{ } & \textbf{Models} & \textbf{2-hop}          & \textbf{3-hop}        & \textbf{4-hop}  \\ 

\midrule
\multirow{3}{*}{\textbf{One-hop}} 
& TF-IDF + BERT \cite{FEVER}    &  57.2   & 49.8    &  45.0     \\
& Oracle + BERT \cite{Oracle}    &   68.3   & 71.5    &  76.4     \\
& CD \cite{CD}    & 81.3   & 80.1  &  78.1  \\
\midrule
\multirow{2}{*}{\textbf{Multi-hop}} 
& Baleen \cite{Baleen}    & 81.2   & \underline{82.5}   &  80.0  \\

& GMR \cite{GMR} & \underline{81.9}   & 82.2   & 
 \underline{80.2}  \\

\midrule
\multirow{2}{*}{\textbf{ }} 
&\textbf{SR-MFV}  & \textbf{82.3}        & \textbf{83.1}        & \textbf{81.1}  \\

\bottomrule
\end{tabular}}

\label{evi-hover}
\end{small}
\end{center}
\end{table}

\subsection{Ablation Study}
To investigate the contribution of each core component in our MFV framework, we conduct a series of ablation experiments on the HoVer development set. As presented in Table~\ref{hover-ablation}, the first group of variants targets the evidence retrieval stage, while the second group focuses on the claim verification stage. 
Specifically, we evaluate the following model variants: 

\textbf{-w/o Reasoning Graph Construction:} It removes the structure-enhanced query generation module. Instead of constructing a reasoning graph, it simply concatenates the claim with previously retrieved evidence to form the next-hop query. 

\textbf{-w/o Multi-hop Retrieval:} It replaces iterative retrieval with a single-hop retrieval strategy, where all evidence sentences are selected at once based on similarity to the claim. 

\textbf{-w/o Subgraph Construction:} It removes the evidence subgraph modeling and instead builds a single unified evidence graph using the claim and all retrieved evidence. This setting allows us to assess the advantage of explicitly modeling reasoning paths.

\textbf{-w/o GraphFormers:} It replaces GraphFormers with a standard GAT for encoding subgraphs. While GAT can model local neighborhood structures, it lacks the global attention scope and structural bias modeling of GraphFormers. 

As illustrated in Table~\ref{hover-ablation}, these ablation results underscore the importance of each component in SR-MFV, showing that progressive retrieval, structure-enhanced query generation, explicit reasoning path modeling collectively underpin its effectiveness in complex multi-hop verification.

\begin{table}[t]
\caption{Performance of ablation study.}
\begin{center}
\begin{small}
\resizebox{0.9\columnwidth}{!}{
\begin{tabular}{llccc}
\toprule
\textbf{ } & \textbf{Models} & \textbf{2-hop} & \textbf{3-hop} & \textbf{4-hop} \\ 
\midrule

\multirow{2}{*}{\textbf{\uppercase\expandafter{\romannumeral1}}} 
& w/o Reasoning Graph Construction & 78.84  &  77.73    & 78.04 \\
& w/o Multi-hop Retrieval & 78.36    & 76.98   &  77.36 \\

\midrule

\multirow{2}{*}{\textbf{\uppercase\expandafter{\romannumeral2}}} 
& w/o Subgraphs Construction & 78.57   & 77.25    & 77.68 \\
& w/o  GraphFormers & 78.63    & 77.47    & 77.79
 \\

\midrule
& \textbf{SR-MFV} & \textbf{79.05}   & \textbf{77.93}   & \textbf{78.36} \\

\bottomrule
\end{tabular}}

\label{hover-ablation}
\end{small}
\end{center}
\end{table}

\subsection{Hyperparameter Sensitivity Analysis}
\begin{figure}[t!]
\centering
    \subfigure[]{\label{Fig-1}
        \begin{tikzpicture}[font=\Large,scale=0.5]
            \begin{axis}[
                legend cell align={left},
                legend style={nodes={scale=1.0, transform shape}},
                xlabel={$Hops$},
                xtick pos=left,
                tick label style={font=\large},
                ylabel style={font=\large},
                ylabel={ },
                ymin=0.70,
                xtick={1,2,3,4},
                xticklabels={$1$, $2$, $3$, $4$},
                legend pos=south east,
                ymajorgrids=true,
                grid style=dashed,
                y tick label style={/pgf/number format/precision=2, /pgf/number format/fixed, /pgf/number format/fixed zerofill},
            ]
            \addplot[
                color=purple,
                dotted,
                mark options={solid},
                mark=diamond*,
                line width=1.5pt,
                mark size=2pt
                ]
                coordinates {
                (1, 0.7850)
                (2, 0.8017)
                (3, 0.7906)
                (4, 0.7803)
                };
            \addlegendentry{FEVER}
            \end{axis}
        \end{tikzpicture}
    }
        \subfigure[]{\label{Fig-2}
        \begin{tikzpicture}[font=\Large,scale=0.5]
            \begin{axis}[
                legend cell align={left},
                legend style={nodes={scale=1.0, transform shape}},
                xlabel={$Hops$},
                xtick pos=left,
                tick label style={font=\large},
                ylabel style={font=\large},
                ylabel={ },
                ymin=0.70,
                xtick={1,2,3,4,5},
                xticklabels={$1$, $2$, $3$, $4$, $5$},
                legend pos=south east,
                ymajorgrids=true,
                grid style=dashed,
                y tick label style={/pgf/number format/precision=2, /pgf/number format/fixed, /pgf/number format/fixed zerofill},
            ]
            \addplot[
                color=purple,
                dotted,
                mark options={solid},
                mark=diamond*,
                line width=1.5pt,
                mark size=2pt
                ]
                coordinates {
                (1, 0.7836)
                (2, 0.7905)
                (3, 0.7867)
                (4, 0.7824)
                (5, 0.7787)
                };
            \addlegendentry{HOVER(hop-2)}
            \end{axis}
        \end{tikzpicture}
    }
        \subfigure[]{\label{Fig-3}
        \begin{tikzpicture}[font=\Large,scale=0.5]
            \begin{axis}[
                legend cell align={left},
                legend style={nodes={scale=1.0, transform shape}},
                xlabel={$Hops$},
                xtick pos=left,
                tick label style={font=\large},
                ylabel style={font=\large},
                ylabel={ },
                ymin=0.70,
                xtick={1,2,3,4,5},
                xticklabels={$1$, $2$, $3$, $4$, $5$},
                legend pos=south east,
                ymajorgrids=true,
                grid style=dashed,
                y tick label style={/pgf/number format/precision=2, /pgf/number format/fixed, /pgf/number format/fixed zerofill},
            ]
            \addplot[
                color=purple,
                dotted,
                mark options={solid},
                mark=diamond*,
                line width=1.5pt,
                mark size=2pt
                ]
                coordinates {
                (1, 0.7534)
                (2, 0.7707)
                (3, 0.7793)
                (4, 0.7746)
                (5, 0.7709)
                };
            \addlegendentry{HOVER(hop-3)}
            \end{axis}
        \end{tikzpicture}
    }
            \subfigure[]{\label{Fig-4}
        \begin{tikzpicture}[font=\Large,scale=0.5]
            \begin{axis}[
                legend cell align={left},
                legend style={nodes={scale=1.0, transform shape}},
                xlabel={$Hops$},
                xtick pos=left,
                tick label style={font=\large},
                ylabel style={font=\large},
                ylabel={ },
                ymin=0.70,
                xtick={1,2,3,4,5},
                xticklabels={$1$, $2$, $3$, $4$, $5$},
                legend pos=south east,
                ymajorgrids=true,
                grid style=dashed,
                y tick label style={/pgf/number format/precision=2, /pgf/number format/fixed, /pgf/number format/fixed zerofill},
            ]
            \addplot[
                color=purple,
                dotted,
                mark options={solid},
                mark=diamond*,
                line width=1.5pt,
                mark size=2pt
                ]
                coordinates {
                (1, 0.7236)
                (2, 0.7427)
                (3, 0.7727)
                (4, 0.7836)
                (5, 0.7796)
                };
            \addlegendentry{HOVER(hop-4)}
            \end{axis}
        \end{tikzpicture}
    }
    
    \caption{Hyperparameter sensitivity analysis(Accuracy vs. $Hops$).}
\label{fig:para}
\end{figure}
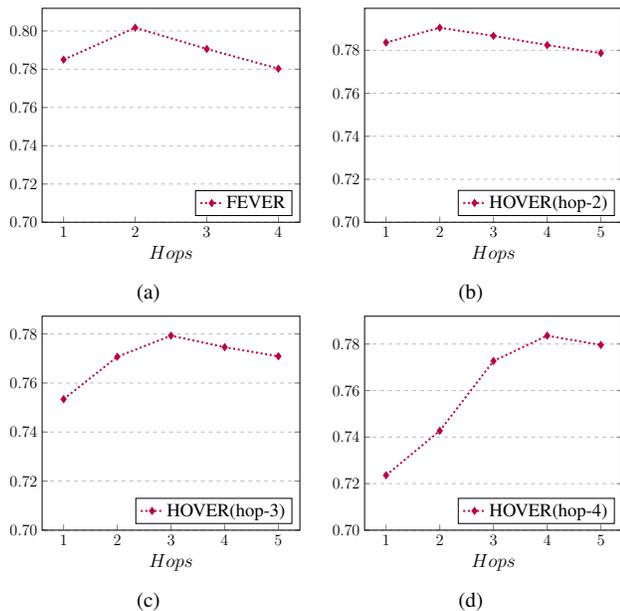

To examine how the number of retrieved evidence hops affects overall verification performance, we conduct the hyperparameter analysis on the FEVER dataset and the three subsets of HoVer. By varying the maximum number of retrieval hops, we aim to understand the trade-off between information sufficiency and noise accumulation across claims of different complexity levels.
As shown in Figure~\ref{fig:para}, the optimal number of retrieval hops varies by dataset. On FEVER, performance peaks at 2 hops, indicating that most claims can be verified with one or two sentences. In contrast, the optimal hop counts for the 2-hop, 3-hop, and 4-hop subsets of HoVer are 2, 3, and 4, respectively—demonstrating that our model adapts well to the reasoning depth required by each claim category.

These results confirm that retrieving too few evidence sentences may lead to insufficient factual grounding, while excessive retrieval introduces redundant or irrelevant information, increasing reasoning difficulty and degrading final verification accuracy. Properly tuning the number of hops is therefore critical for balancing completeness and conciseness in multi-hop fact verification.

\section{Conclusion}
In this paper, we present a structural reasoning framework for multi-hop fact verification that explicitly models the reasoning path across both evidence retrieval and claim verification stages. 
We construct reasoning graphs during retrieval to generate structure-enhanced queries, which guide the acquisition of evidence along a coherent reasoning path. 
In the verification phase, a sequence of evidence subgraphs is progressively built to explicitly model the inference process, enabling the system to capture the evolving nature of multi-hop reasoning—an essential capability for verifying complex claims.
Experimental results on the FEVER and HoVer datasets demonstrate that our method outperforms strong baselines in both retrieval precision and verification accuracy, particularly on complex claims requiring long-range evidence coordination.



\begin{thebibliography}{00}

\bibitem{FEVER} J. Thorne, A. Vlachos, C. Christodoulopoulos, and A. Mittal, “FEVER: a large-scale dataset for fact extraction and verification,” in Proc. Conf. North Amer. Chapter Assoc. Comput. Linguistics: Human Language Technol. (NAACL-HLT), 2018, pp. 809–819.

\bibitem{HOVER} Y. Jiang, S. Bordia, Z. Zhong, C. Dognin, M. K. Singh, and M. Bansal, “HoVer: a dataset for many-hop fact extraction and claim verification,” in Findings of the Assoc. Comput. Linguistics: EMNLP, 2020, pp. 3441–3460.

\bibitem{GEAR} 	J. Zhou, X. Han, C. Yang, Z. Liu, L. Wang, C. Li, and M. Sun, “GEAR: graph-based evidence aggregating and reasoning for fact verification,” in Proc. Annu. Meeting Assoc. Comput. Linguistics (ACL), vol. 1, 2019, pp. 892–901.

\bibitem{KGAT} Z. Liu, C. Xiong, M. Sun, and Z. Liu, “Fine-grained fact verification with kernel graph attention network,” in Proc. Annu. Meeting Assoc. Comput. Linguistics (ACL), 2020, pp. 7342–7351.
 
\bibitem{DREAM} W. Zhong, J. Xu, D. Tang, Z. Xu, N. Duan, M. Zhou, J. Wang, and J. Yin, “Reasoning over semantic-level graph for fact checking,” in Proc. Annu. Meeting Assoc. Comput. Linguistics (ACL), 2020, pp. 6170–6180.

\bibitem{EvidenceNet} Z. Chen, S. C. Hui, F. Zhuang, L. Liao, F. Li, M. Jia, and J. Li, “EvidenceNet: evidence fusion network for fact verification,” in Proc. Int. World Wide Web Conf. (WWW), 2022, pp. 2636–2645.

\bibitem{HESM} S. Subramanian and K. Lee, “Hierarchical evidence set modeling for automated fact extraction and verification,” in Proc. Conf. Empirical Methods Natural Language Process. (EMNLP), vol. 1, 2020, pp. 7798–7809.

\bibitem{CO-GAT} Y. Lan, Z. Liu, Y. Gu, X. Yi, X. Li, L. Yang, and G. Yu, “Multi-evidence based fact verification via a confidential graph neural network,” IEEE Trans. Big Data, vol. 11, 2025, pp. 426–437.

\bibitem{DeBERTa} P. He, X. Liu, J. Gao, and W. Chen, “DeBERTa: decoding-enhanced BERT with disentangled attention,” in Proc. Int. Conf. Learn. Represent. (ICLR), 2021.


\bibitem{SAGP} J. Si, Y. Zhu, and D. Zhou, “Exploring faithful rationale for multi-hop fact verification via salience-aware graph learning,” in Proc. AAAI Conf. Artif. Intell. (AAAI), 2023, pp. 13573–13581.

\bibitem{GERE} J. Chen, R. Zhang, J. Guo, Y. Fan, and X. Cheng, “GERE: generative evidence retrieval for fact verification,” in Proc. Int. ACM SIGIR Conf. Res. Develop. Inf. Retrieval (SIGIR), 2022, pp. 2184–2189.

\bibitem{TARSA}  J. Si, D. Zhou, T. Li, X. Shi, and Y. He, “Topic-aware evidence reasoning and stance-aware aggregation for fact verification,” in Proc. Annu. Meeting Assoc. Comput. Linguistics / Int. Joint Conf. Natural Language Process. (ACL/IJCNLP), vol. 1, 2021, pp. 1612–1622.

\bibitem{Proofver} 	A. Krishna, S. Riedel, and A. Vlachos, “ProoFVer: natural logic theorem proving for fact verification,” Trans. Assoc. Comput. Linguistics, vol. 10, pp. 1013–1030, 2022.

\bibitem{UNCNLP} Y. Nie, H. Chen, and M. Bansal, “Combining fact extraction and verification with neural semantic matching networks,” in Proc. AAAI Conf. Artif. Intell. (AAAI), 2019, pp. 6859–6866.

\bibitem{MLA} C. Kruengkrai, J. Yamagishi, and X. Wang, “A multi-level attention model for evidence-based fact checking,” in Findings Assoc. Comput. Linguistics: ACL/IJCNLP, 2021, pp. 2447–2460.

\bibitem{ICMI} 	H. Wang, Y. Li, Z. Huang, and Y. Dou, “IMCI: integrate multi-view contextual information for fact extraction and verification,” in Proc. Int. Conf. Comput. Linguistics (COLING), 2022, pp. 1412–1421.

\bibitem{Oracle} A. Hanselowski, H. Zhang, Z. Li, D. Sorokin, B. Schiller, C. Schulz, and I. Gurevych, “UKP-Athene: multi-sentence textual entailment for claim verification,” in Proc. Fact Extraction Verification Workshop (FEVER)@EMNLP, 2018, pp. 103–108.

\bibitem{CD} M. Fajcik, P. Motlícek, and P. Smrz, “Claim-Dissector: an interpretable fact-checking system with joint re-ranking and veracity prediction,” in Findings Assoc. Comput. Linguistics: ACL, 2023, pp. 10184–10205.

\bibitem{Baleen} O. Khattab, C. Potts, and M. A. Zaharia, “Baleen: robust multi-hop reasoning at scale via condensed retrieval,” in Adv. Neural Inf. Process. Syst. (NeurIPS), 2021, pp. 27670–27682.

\bibitem{GMR} O. Khattab, C. Potts, and M. A. Zaharia, “Baleen: robust multi-hop reasoning at scale via condensed retrieval,” in Adv. Neural Inf. Process. Syst. (NeurIPS), 2021, pp. 27670–27682.

\bibitem{DeFormer} Q. Cao, H. Trivedi, A. Balasubramanian, and N. Balasubramanian, “DeFormer: Decomposing pre-trained transformers for faster question answering,” in Proc. Annu. Meeting Assoc. Comput. Linguistics (ACL), 2020, pp. 4487–4497.

\bibitem{ListT5} K. Jiang, R. Pradeep, and J. Lin, “Exploring listwise evidence reasoning with T5 for fact verification,” in Proc. Annu. Meeting Assoc. Comput. Linguistics / Int. Joint Conf. Natural Language Process. (ACL/IJCNLP), vol. 2, 2021, pp. 402–410.

\bibitem{SURVEY} Z. Guo, M. S. Schlichtkrull, and A. Vlachos, “A survey on automated fact-checking,” Trans. Assoc. Comput. Linguistics, vol. 10, pp. 178–206, 2022.

\bibitem{AER} N. De Cao, G. Izacard, S. Riedel, and F. Petroni, “Autoregressive entity retrieval,” in Proc. Int. Conf. Learn. Represent. (ICLR), 2021.

\bibitem{Politihop} W. Ostrowski, A. Arora, P. Atanasova, and I. Augenstein, “Multi-hop fact checking of political claims,” in Proc. Int. Joint Conf. Artif. Intell. (IJCAI), 2021, pp. 3892–3898.

\bibitem{Self-RAG} A. Asai, Z. Wu, Y. Wang, A. Sil, and H. Hajishirzi, “Self-RAG: Learning to retrieve, generate, and critique through self-reflection,” in Proc. Int. Conf. Learn. Represent. (ICLR), 2024.

\bibitem{faithful-rationale} J. Si, Y. Zhu, and D. Zhou, “Exploring faithful rationale for multi-hop fact verification via salience-aware graph learning,” in Proc. AAAI Conf. Artif. Intell. (AAAI), 2023, pp. 13573–13581.

\bibitem{Multi-Hop-DR} W. Xiong, X. L. Li, S. Iyer, J. Du, P. Lewis, W. Y. Wang, Y. Mehdad, S. Yih, S. Riedel, D. Kiela, and B. Oguz, “Answering complex open-domain questions with multi-hop dense retrieval,” in Proc. Int. Conf. Learn. Represent. (ICLR), 2021.

\bibitem{RMHR} O. Khattab, C. Potts, and M. A. Zaharia, “Baleen: Robust multi-hop reasoning at scale via condensed retrieval,” in Adv. Neural Inf. Process. Syst. (NeurIPS), 2021, pp. 27670–27682.

\bibitem{RECOMP} F. Xu, W. Shi, and E. Choi, “RECOMP: Improving retrieval-augmented LMs with context compression and selective augmentation,” in Proc. Int. Conf. Learn. Represent. (ICLR), 2024.

\bibitem{Causal-Walk} C. Zhang, L. Zhang, and D. Zhou, “Causal Walk: Debiasing multi-hop fact verification with front-door adjustment,” in Proc. AAAI Conf. Artif. Intell. (AAAI), 2024, pp. 19533–19541.

\bibitem{reread} X. Hu, Z. Hong, Z. Guo, L. Wen, and P. S. Yu, “Read it twice: Towards faithfully interpretable fact verification by revisiting evidence,” in Proc. Int. ACM SIGIR Conf. Res. Develop. Inf. Retrieval (SIGIR), 2023, pp. 2319–2323.

\bibitem{LOREN} J. Chen, Q. Bao, C. Sun, X. Zhang, J. Chen, H. Zhou, Y. Xiao, and L. Li, “LOREN: Logic-regularized reasoning for interpretable fact verification,” in *Proc. AAAI Conf. Artif. Intell. (AAAI)*, 2022, pp. 10482–10491.


\bibitem{GraphFormers} J. Yang, Z. Liu, S. Xiao, C. Li, D. Lian, S. Agrawal, A. Singh, G. Sun, and X. Xie, “GraphFormers: GNN-nested transformers for representation learning on textual graph,” in *Adv. Neural Inf. Process. Syst. (NeurIPS)*, 2021, pp. 28798–28810.

\bibitem{MRR-FV} L. Zheng, C. Li, L. Zhang, H. Jia, S. Wang, Z. Liu, and X. Zhang, “MRR-FV: Unlocking Complex Fact Verification with Multi-hop Retrieval and Reasoning,” in *Proc. AAAI Conf. Artif. Intell. (AAAI)*, 2025, vol. 39, no. 24, pp. 26066–26074.

\bibitem{RAV} L. Zheng, C. Li, X. Zhang, Y. Shang, F. Huang, and H. Jia, “Evidence Retrieval is almost All You Need for Fact Verification,” in *Findings of the Association for Computational Linguistics (ACL)*, 2024, pp. 9274–9281.







\end{thebibliography}
\end{document}